\documentclass[journal]{IEEEtran}

\usepackage{times}
\usepackage{epsfig}
\usepackage{graphicx}
\usepackage{amsmath}
\usepackage{amssymb}

\usepackage{multirow}
\usepackage[skip=10pt]{caption}
\usepackage{booktabs}

\usepackage{times}
\usepackage{epsfig}
\usepackage{graphicx}
\usepackage{amsmath}
\usepackage{amssymb}
\usepackage{color}
\usepackage[normalem]{ulem}
\usepackage{cite}

\usepackage{algorithm}
\usepackage{algorithmic}
\usepackage[algo2e]{algorithm2e} 

\usepackage{mathtools}

\begin{document}

%
\title{Biased Mixtures Of Experts: Enabling Computer Vision Inference Under
Data Transfer Limitations}
%
%
%

\author{Alhabib Abbas and
        Yiannis Andreopoulos

\thanks{The authors are researchers in the Electronic and Electrical Engineering Department of
University College London, London, UK, Roberts Building,  WC1E 7JE (e-mail:
\{alhabib.abbas.13, i.andreopoulos\}@ucl.ac.uk). The authors acknowledge support from the UK EPSRC grants EP/R025290/1 and EP/P02243X/1.}}
%
%

\markboth{IEEE\ TRANSACTIONS\ on Image processing, To Appear}%
{Shell \MakeLowercase{\textit{et al.}}: Bare Demo of IEEEtran.cls for IEEE Journals}
%



\maketitle


\begin{abstract}

We propose a novel mixture-of-experts class to optimize computer vision
models in accordance with data transfer limitations at test time. Our approach
postulates
that the minimum acceptable amount of data allowing for highly-accurate results
  can vary for different   input space  partitions.  Therefore, we consider
 mixtures  where experts require different amounts of data, and  train  a
sparse gating function to divide the input space for each expert. By appropriate
hyperparameter selection, our approach is able to bias  mixtures of experts
 towards selecting specific experts over others.
In this way, we show that the data transfer optimization between visual sensing
and processing can be solved as a convex optimization problem.   To demonstrate
 the
relation between  data availability and  performance, we evaluate biased
mixtures on a range of mainstream computer vision problems,
namely:  (i) single shot detection, (ii)  image super resolution, and (iii)
realtime
video action classification.  For all cases, and when experts constitute
modified baselines to meet different limits on allowed data utility, biased
mixtures significantly outperform 
previous work optimized to
meet the same constraints on available data.
\end{abstract}


\vspace*{-3mm}
\section{Introduction}
\label{sec:intro}
%
%

When enough data is provided at test time, deep neural
networks   perform  well  for a wide range of challenging computer vision
tasks. This is true especially for large models, as it is now well understood
that the performance of neural networks scales with the number of  trainable
weights  and the   dimensionality of inputs processed during inference \cite{howard2017mobilenets,huang2017speed}.
However, the precondition of data availability at test time is only possible when visual sensors and learned inference models coexist in hardware, which excludes cases where data is collected from sensors to be transferred and processed in remote environments.
To bridge the gap between the input requirements of models that exist in such contexts, it is important to
design  models that can perform well when available communication resources
are limited between the visual
sensing and neural network processing parts of the system.  
%
\begin{figure}[h] \centering \centering \hspace*{-4mm} \vspace*{-2.3mm}\includegraphics[scale=0.30]{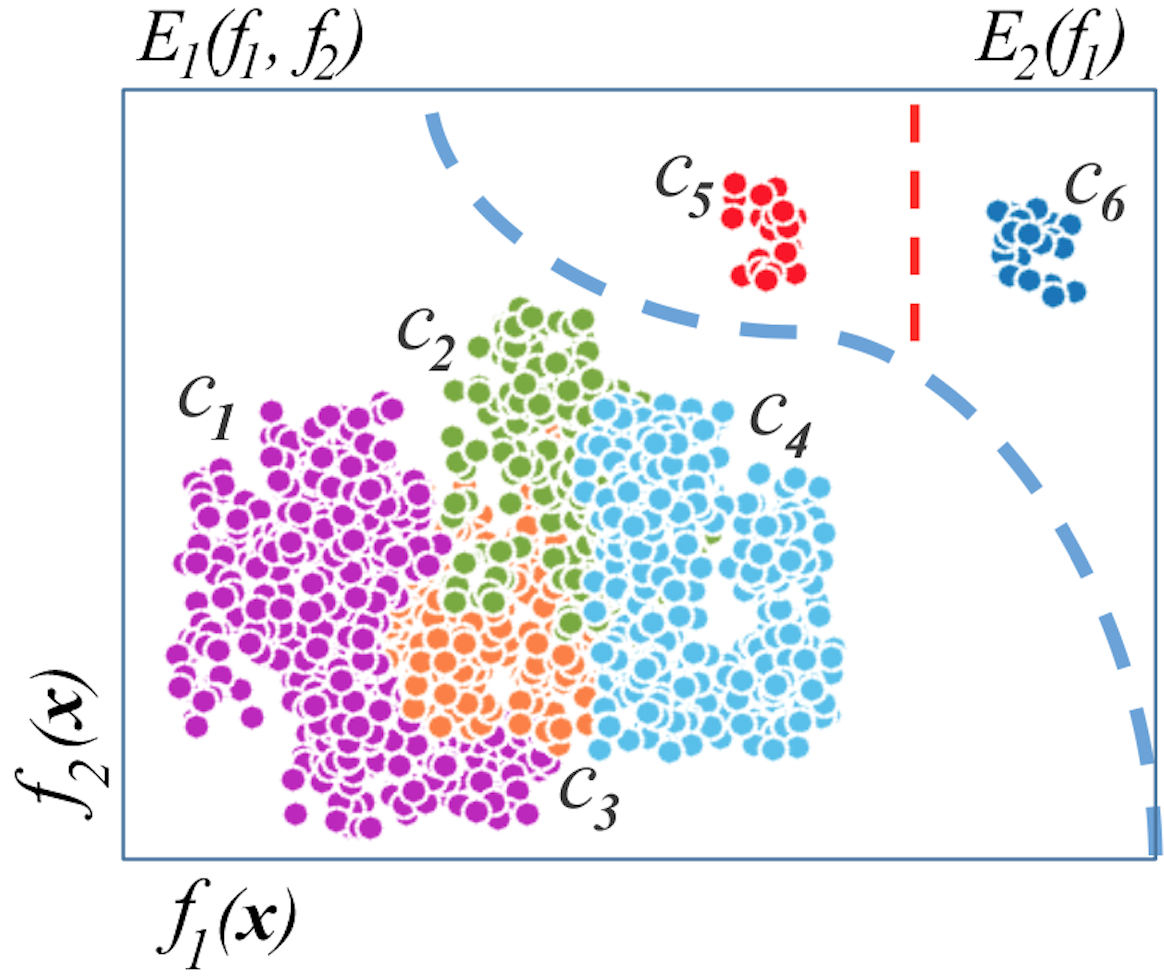}
          \caption{\label{fig:space_partitions_example} Sample space of a
classification task using two features $f_1(\boldsymbol x)$ and $f_2(\boldsymbol x)$ of arbitrary inputs $\boldsymbol x$,
where colours  indicate different classes $c_1-$$c_6$. The  blue line shows an instance
of  a learnable input space partition $E_1(f_1, f_2)$, and the red line shows a one-dimensional
classification boundary learnable by a designated expert $E_2(f_1)$ with
reduced data requirement relative to $E_1(f_1, f_2)$.}

\vspace*{-3mm} \end{figure}
For instance,  cloud-based visual analysis,   remote medical imaging, low-latency game streaming services, and drone or Internet-of-Things oriented computer vision   \cite{chuah2016layered}\cite{zhaoyang2017tsd,
srivastava2015superresolution,martin2013characterizing},  have stringent
constraints on the amount of data that can be provided between data-producing
clients and data-consuming models on cloud  servers.  
In order to bring  computer vision models to wider practical use, it is therefore imperative to provide a solution to  data availability constraints at test time. 

Since deep learning models typically require a fixed amount of data for inference regardless of the specific nature of inputs to process, this leads to unnecessary and often unachievable demands
in the amount of required   data traffic for remote inference.
Although some work has been devoted to input dimensionality reduction \cite{makhzani2015adversarial,hinton2006reducing,wang2014generalized}
and rate-constrained model optimization for specific 
 tasks  \cite{zhaoyang2017tsd,jubran2018rate}, to the best of our knowledge,
no  task-agnostic method has been proposed that explicitly addresses   data
scarcity
at test time by considering the variance between
different domains in input space. The example of Figure \ref{fig:space_partitions_example}
illustrates
a classification task where the acceptable data cost of inference can vary
for different input space
partitions. That is,  two features $f_1$ and $f_2$   can
be used to  classify
the bottom-left examples in Figure \ref{fig:space_partitions_example}, while
one feature $f_1$ suffices for distinguishing
class $c_5$ examples from class $c_6$ examples on the top-right. Reducing the retained
dimensions directly correlates with the \textit{data cost} of inference.
To leverage  inherent variances across different input space partitions,
and by selecting among two experts $E_1$ and $E_2$ which
respectively require $d_1$ and $d_2$ bytes per input where $d_1>d_2$,
    decision boundaries can be determined to appropriately pass more data
for more difficult inputs.    Learning decision boundaries similar to those
of Figure
\ref{fig:space_partitions_example}  can allow sensors to remotely communicate
data as necessary,
subject to the general position of an input within its respective space.
This  reduces the overall data cost of  inference that is accurate enough
for the task at hand. Consequentially,
this in turn can  relieve unnecessary load on communication resources that exist
between sensors and remote machines used for visual inference. Our work proposes
a solution to learning such decision boundaries directly from data for any
set of pretrained experts, and for
any specified limit on  data cost.
Our contributions are summarised below:


\vspace*{-1mm}

\begin{enumerate}

\item We introduce a novel    class of mixtures-of-experts, wherein some
experts
are favored to others by design.  When experts of different data requirements
are included, this allows mixtures to  meet different
constraints on allowed data utility.

\item We propose two methods to train biased mixtures such that
 input space is effectively partitioned for each expert to realize data-efficient
mixtures. 
\item We show that  data transfer optimization between visual sensing and
processing can be formulated as a convex optimization problem, and present
an ablation study of the benefit of  biased mixtures under different contexts
of allowed limits on  data  utility. 

\end{enumerate}

The expert utility biasing method proposed in this paper  can be applied
to reduce the data cost  of any model wherein the size of inputs can be subsampled
or reduced.
To illustrate this, we train and  validate  on a variety of tasks spanning
multiple domains. Specifically, we validate on the tasks of: single shot
object detection
from
the work of Wei  \textit{et. al} \cite{liu2016ssd},     realtime video action
classification from the
work of Zhang \textit{et. al} in \cite{zhang2018real} and Chadha \textit{et.
al} \cite{chadha2017video}, and image super resolution from the work of Shi
\textit{et. al} \cite{shi2016real} and Dong \textit{et. al} \cite{dong2016accelerating}.
The remainder of this paper is organized as follows: In Section
\ref{sec:related_work},
we give an overview of recent work on rate and complexity optimization. Section
 \ref{sec:biased_expert_selection} details  the proposed biased expert selection
 and describes its general architecture and how it is trained. In
Section \ref{sec:evaluation}
we evaluate the performance of the proposed method on all tasks, and illustrate
the benefits that biased mixtures of experts can provide on multiple models
for each task. Finally, Section \ref{sec:conclusion} summarises our findings
and
outlines   possible directions for future work.

%

 \vspace*{-2mm}

\section{Related Work \\ \label{sec:related_work}}

\vspace*{-1mm}

Within the field of compact image representation, and in order to communicate
data-efficient codes across networks for remote processing,
 directly engineered compression techniques were extensively studied to culminate
in existing image compression standards \cite{miano1999compressed,roelofs1999png}.
More recently, learned methods \cite{minnen2018joint,van2016conditional,wang2014generalized}
have attracted
 attention as the next step towards more data-driven image compression. Salient among recent advances in this domain
are variational autoencoders \cite{balle2016end,mescheder2017adversarial,rolfe2016discrete}
   and adversarial models\cite{radford2015unsupervised,goodfellow2014generative,denton2015deep}.
In order to adapt learned codes to arithmetic coders, state-of-the-art proposals
on learned compression \cite{van2016conditional,oord2016pixel,salimans2017pixelcnn++,mentzer2019practical}
additionally learn context models to predict posteriors of latent code components
conditional on all preceding components.
Specifically, and to move learned compression closer to replacing established
 coders\cite{miano1999compressed,roelofs1999png}, context models  \cite{mentzer2019practical,salimans2017pixelcnn++}
\ use tractable masked
convolutions to regulate entropies of   obtained image representations such
that they can be coded more effectively by subsequent
entropy coders. In distributed systems of visual analysis,  and in order
to reduce throughput requirements
on input,  latent states of learned image reconstruction machines
 \cite{balle2016end,mescheder2017adversarial,radford2015unsupervised,goodfellow2014generative}
and entropy regulated compressors \cite{mentzer2019practical,salimans2017pixelcnn++,minnen2018joint,van2016conditional}
can
be used instead of full-length inputs as representative signals  to remote
inference
models.

Other studies consider the regulation of input volumes for \textit{complexity} optimization,  and
propose modifications that are applicable to a wide range of models. In this
realm, proposals such as static model pruning \cite{han2015deep,han2015learning,howard2017mobilenets},  reduce complexity by  modifying
models in a persistent manner for all inputs at test time. More recent  proposals \cite{bengio2015conditional,bengio2013estimating,lin2017runtime, shazeer2017outrageously}
   show how the test-time complexity of
very large networks can be substantially reduced by conditioning computation to the content of feature maps at runtime, and do so by training
 external agents to enable or disable  different parts of  models subject to the unique properties of each input.    However, all of the aforementioned works optimize solely for complexity, and always consider  the maximum amount of input to be available
at test time.
Other proposals  also studied specific vision
tasks in order to reduce the data requirement  of
deep neural  network models.  For example, this can be seen
in previous  work  \cite{zhaoyang2017tsd,zhang2018real,chadha2017video},
where  input volumes are reduced by distilling input\ sequences to their
most useful elements  before relaying  to remote
servers for semantic analysis. Other  work  \cite{li2019learning,wong2017low}
  mainly focused
on   task-specific mappings of inputs  onto lower-dimensional space before
 training    with more data-efficient models, and recent advances in domain
adaptation and transfer learning \cite{tzeng2017adversarial,sankaranarayanan2018generate,long2018conditional}
 can also be used to learn compressed codes tuned to particular models. However,
for any specified source distribution, domain adaptation  \cite{tzeng2017adversarial,sankaranarayanan2018generate,long2018conditional}
and other proposals mentioned above \cite{zhaoyang2017tsd,zhang2018real,chadha2017video}
equally compact all sampled inputs to fixed length codes, and varying degrees
of entropy  among input examples are ignored. 
As such, low-entropy inputs (which contain less information relative to others) are  mapped to redundantly long code-lengths, and  subsequently incur unnecessary loads on data transfer  assets and inference complexity.  In this sense, while the aforementioned advances are important in determining
useful transformations to  fixed-length codes, complementary techniques
are necessary to determine required code lengths
prior to compression and inference.

\begin{figure*}[ht!] \centering \centering \hspace*{-3.mm} \vspace*{-1.5mm}
\includegraphics[scale=0.54]{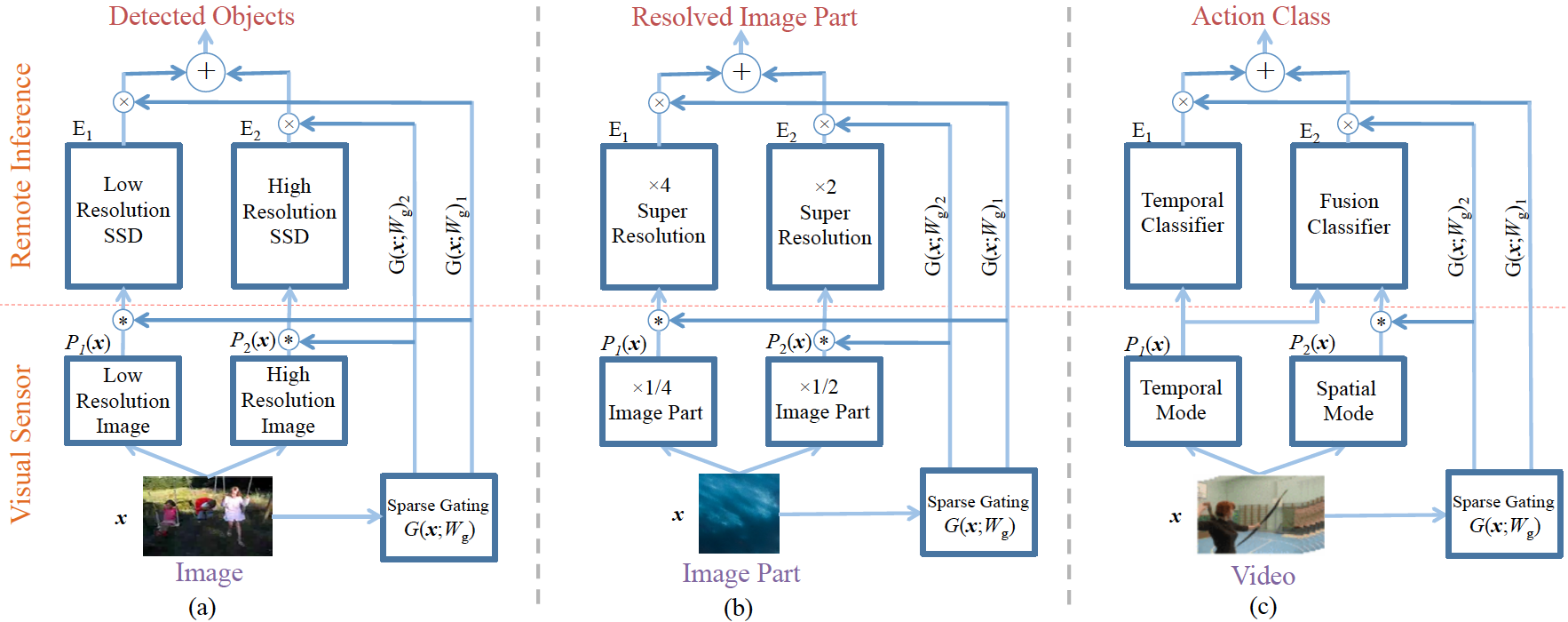}
          \caption{\label{fig:bmoe_illustration} An illustration of how biased
mixtures of experts can be applied for different computer vision tasks.  ($*$)
is a special operator that transmits
data to  remote inference parts of the model whenever it receives a non-zero
gate value. From
left to right: (a)  single shot detection (SSD), (b)  image super resolution,
and
(c) realtime action classification.\textcolor[rgb]{0,0.501961,1}{}    
\textbf{\textcolor{red}{}}} \vspace*{-2mm} \end{figure*}

In our work, we  consider the \textit{data cost} optimization
problem in a task-agnostic manner, and determine required input volumes  \textit{prior
to visual inference}.  Specifically, we consider how  input space partitions
 vary in the amount of  data required per input in order to ensure good performance,
and  leverage this variance to train more data-efficient
mixtures of experts. To do so, we take inspiration from recent work  \cite{shazeer2017outrageously,lin2017runtime,howard2017mobilenets}
to propose a  mixture of experts where expert utility is biased towards specific
experts. While meeting
predefined constraints on expert utility bias, we train a  sparse gating
function to select the most adequate
expert to use from a set of experts of varied input requirements. Importantly,
our  method does not modify any pre-existing methods for complexity optimization
or \textcolor[rgb]{0,0,0}{ task-specific} data cost  reduction. As such, our proposal can be applied
in conjunction with recent proposals on learned compression \cite{oord2016pixel,salimans2017pixelcnn++,mentzer2019practical}
and  domain adaptation \cite{tzeng2017adversarial,sankaranarayanan2018generate,long2018conditional}
 to reduce
the data cost of  visual inference.  We show that
our method can be augmented   in accordance with any set of pre-trained experts to partition
input space such that constraints on data availability are met at test time, while providing the best  possible accuracy of inference.

\section{Biased Expert Selection\\  \label{sec:biased_expert_selection}}

\subsection{General Architecture  Formulation} \label{sec:highly_compressed_bitstreams}
\vspace*{-0.1mm}

  Let $\mathcal{E}$ denote a 
mixture of $N$ experts where  $\mathcal{E}=\{E_1,E_2,
..., E_N\} $, and each  expert $E_n$ is a  modified variant of a task-performing
baseline   model. Per  input    $\boldsymbol x$,   a   gating function determines
the contribution of the $n^{th}$ expert as:
 
\begin{equation}
\label{eq:softmax}
G(\boldsymbol x; \mathcal{W}_g)_n = \frac{e^{f(\boldsymbol x;\mathcal{W}_g)_n}}{\sum^{N}_{m\neq n} e^{f(\boldsymbol x;\mathcal{W}_g)_m}}
\end{equation} where $\mathcal{W}_g$ is a set of trainable weight parameters,
$m$ denotes remaining gate indices,
and $f(\boldsymbol x;\mathcal{W}_g)\in  \mathbb{R}^{N}$ is the output of
a specified gating model (e.g, a multi-layer perceptron). The output $\boldsymbol
y$\
of the mixture is: 
\begin{equation}\label{eq:moe_output}
\begin{aligned}
\boldsymbol y=\sum_{n=1}^{N} G(\boldsymbol x; \mathcal{W}_g)_n  E_n(P_n(\boldsymbol x))
\end{aligned}
\end{equation}where $ P_n$ is a preprocessing function to accommodate $\boldsymbol
x$ for the $n^{th}$ expert (e.g., $ P_n$ performs subsampling   if  $E_n$
 ingests sub-sampled inputs). Mixtures-of-experts are typically trained using
a task  loss
that  calculates
the error  between a provisioned ground-truth   and $\boldsymbol y$. 
In our proposed Biased Mixtures-of-Experts (BMoE) paradigm,    experts  are  activated
only when needed, and activating some experts is more favorable to activating
others.  In addition, all experts
are optimized before training the mixture, and    loss functions of $\boldsymbol y$ are back-propagated
through the gating function exclusively. In Figure \ref{fig:bmoe_illustration}
we illustrate some examples
of how biased mixtures can be applied  to different  tasks.

To adjust  mixtures for biased expert selection,  we denote the desired amount
of bias in expert selection by  $ \boldsymbol b$, where each of its components $b_n$
specifies per batch the ratio of input examples to pass to each $n^{th}$
expert. Importantly, elements of $\boldsymbol b$ denote \textit{frequencies} of
use as ratios and cannot be assigned negative values   (e.g., setting $ b_n=0.1$
to use expert $E_n$ 10\% of the time),  giving the  properties $0\leq b_n \leq 1$ and $||\boldsymbol b||_1 = 1$. We consider two methods of training
for biased expert selection: \textit{(i)}      a soft regularization
approach where a regularization term  is included in the total loss to encourage
bias, and \textit{(ii)} fixing the  average data cost \textit{per batch},
by enforcing a constant number of training examples to each expert in accordance
with $\boldsymbol b$ and training only with respect to the task loss. Both
methods encourage mixtures of experts to maximize
performance while meeting the specified bias, and we describe in detail each
method in the following:

\subsection{Soft Bias Regularization\\ }

When using soft bias regularization, the most suitable expert to use is selected
 \textit{per input} via a sparse gating function, and all  other experts
are omitted. To do so, akin to \cite{shazeer2017outrageously}   for each input $\boldsymbol x$ only
the expert associated with the highest gate value is considered for inference,
and we write the  sparse gating function
as:
 \vspace{-10pt}
 
\begin{equation}\label{eq:sparse_G}
\begin{aligned}
G(\boldsymbol I;\mathcal{W}_g)_{n}=\psi\text{}(f(\boldsymbol x \mathcal{;W}_g))_{n}
\cdot \frac{e^{f(\boldsymbol x;\mathcal{W}_g)_n}}{\sum^{N}_{m\neq n}e^{f(\boldsymbol x;\mathcal{W}_g)_m}} \;  
\end{aligned}
\end{equation}
where $\psi( f(\boldsymbol x \mathcal{;W}_g))$ is a non-linear operator which returns a one-hot vector indicating the top value in $f(\boldsymbol I \mathcal{;W}_g)$. From (\ref{eq:sparse_G}) we   also define the utility of each $n^{th}$ expert
$ u_n 
$ as its  total contribution  per batch $\mathcal{X}$ comprising $M$ examples:

\begin{equation}\label{eq:utility}
\begin{aligned}
 u_n= \frac{1}{M}\sum_{\boldsymbol x \in  \mathcal{X}} G(\boldsymbol
x; \mathcal{W}_g)_{n} \;
\end{aligned}
\end{equation}  
and we calculate the bias regularization loss 
$l_\text{bias}$  as a function of  $\boldsymbol u$ and
the specified  bias vector  $\boldsymbol b$: 
\begin{equation}\label{eq:loss_bias}
\begin{aligned}
 l_{\text{bias}} =  -w_{\text{bias}} \; log(1- \frac{1}{\sqrt{2}} 
 {||\boldsymbol u-\boldsymbol b||}_{2})
\end{aligned}
\end{equation}where $w_{\text{bias}}$ is  a hyperparameter  to control
the amount of   bias to impose on the mixture. Since $\boldsymbol u$ and $\boldsymbol b$  describe
frequencies as ratios and  $||\boldsymbol u||_1=||\boldsymbol b||_1 = 1$, the distance $||\boldsymbol u-\boldsymbol b||_2$  is normalized
by   $\sqrt{2}$    to ensure the expression within the log function is always
positive ($\sqrt{2}$     is the maximum possible euclidian distance between vectors
with  an $L_1$ norm of one). By applying the modifications to the gating function in (\ref{eq:sparse_G}),
and  including the bias regularization loss in
(\ref{eq:loss_bias})  to the total loss, the mixture of experts is simultaneously
trained
to   maximize task performance and meet the specified  bias.


\subsection{Batchwise Bias Enforcement \\ \label{sec:combined_loss}}

\begin{figure}[h] \centering \centering \hspace*{-2.mm} \includegraphics[scale=0.3]{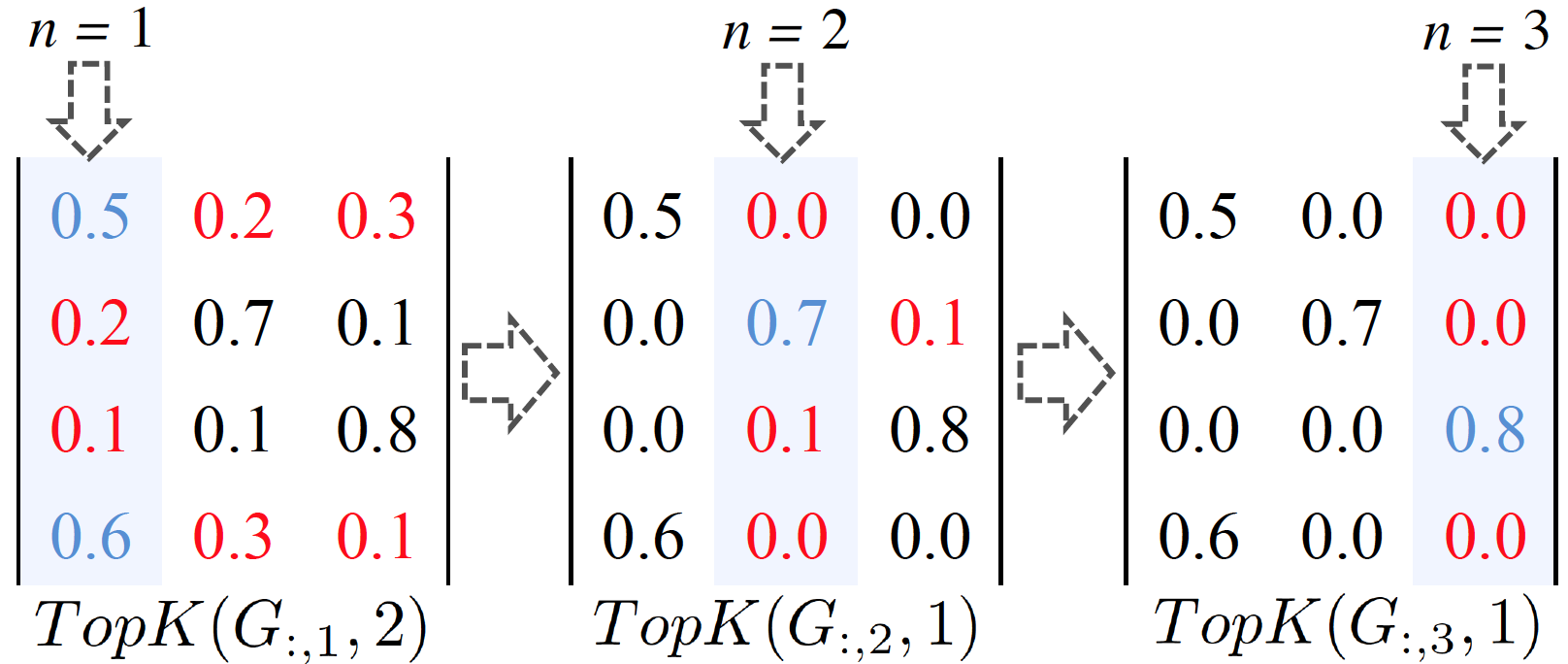}
\vspace{-2pt} \caption{ \label{fig:bias_enforcement} Batchwise bias enforcement
example when $N=3$, $M=4$ and $\boldsymbol b=[0.50,0.25,0.25]$. Inputs  are selected
per batch by iteratively sorting
and selecting the top $M b_n$ highest gate values for each $n^{th}$
expert. Gates subsequently set
to zero are highlighted in red, and  top $(Mb_n)$ values are highlighted in blue.} \vspace{-14pt}  \end{figure}

In our second proposal, rather than encourage  mixtures to align the utility
of their  experts with the specified bias, we enforce bias \textit{per  batch}
in accordance with $\boldsymbol b,$ and train the mixture  only with respect
to its task loss. This in effect trains mixtures to make better expert
selections for each input, while meeting the bias constraint for every batch.
Specifically, with a batch size of $M$,  batches are segmented such that
 $M b_n$ examples are passed to  each $n^{th}$ expert.
To do so,   starting from (\ref{eq:softmax}), we  consider $G\in\mathbb{R}^{M\times
N}$ as an $M$ sized  batch of gate vectors $G(\boldsymbol x; \mathcal{W}_g)$,
and perform the procedure described in  Algorithm \ref{alg:batchwise_bias_enforcement}.
For each $n^{th}$ expert, we denote gate values assigned to columns of input
 as $G_{:,n}$ and illustrate this in Figure \ref{fig:bias_enforcement}.

%
 \begin{algorithm}[H]
 
 \caption{Batchwise Bias Enforcement}
 \label{alg:batchwise_bias_enforcement}
 \begin{algorithmic}[1]
 \renewcommand{\algorithmicrequire}{\textbf{Input:}}
 \renewcommand{\algorithmicensure}{\textbf{Output:}}
 \REQUIRE Soft gates batch  $G\in \mathbb{R}^{M\times N}$

  \FOR {$n =  1$ to $n=N$}
  \STATE   
  $K \gets M\boldsymbol b_n$ \\
  Calculate number of  inputs to pass to the $n^{th}$ expert 

  \STATE    $T\gets$ $TopK(G_{:,n},K)$ \\Find  top $K$ values corresponding
to  the $n^{th}$ expert

 \FOR {$i =  1$ to $i=M$}

   \IF {$T_{i}$ $\neq$ $0$ }

\STATE $G_{i,j} \gets 0 \;\;\forall j \neq n$ \\ 
 For
the $i^{th}$ input, set all   gate values  not corresponding to $n^{th}$
expert to $0$      
\ELSE
\STATE  $G_{i,n} \gets 0 \;\;$ \\
Set gate value corresponding to the   $i^{th}$ input  and $n^{th}$ expert
to $0$ 
  \ENDIF
  \ENDFOR
   \ENDFOR    $_{\text{}}  $
 \end{algorithmic}

 \end{algorithm}

%
%
%
%
%

\subsection{Selecting Bias Values for Data  Cost Optimization}

So far, we  discussed how  biased mixtures are trained to  make informed expert selections
when a  bias vector $\boldsymbol b$ specifies  the frequency of expert utility.  Here we  detail our method for selecting useful biases that can optimize
performance under
different constraints on data utility. We consider the inference data cost
 vector $\boldsymbol d$, where  each of its components
$d_n$  is the size of input
volumes per example as seen by each expert
(i.e., the data cost associated with $P_n(\boldsymbol
x)$). When mixtures are biased, and an ample number of samples is considered,
 the average data cost is then expressed as $\bar d=\boldsymbol b\boldsymbol
d^\mathrm{T}=\sum_{n=1}^N
 b_n d_n$. In this way, the biasing vector $\boldsymbol b$ can
be tuned  to allow for different average data costs of inference in the interval
$[d_{min},d_{max}]$, where $d_{min}$ and $d_{max}$  are the minimum and maximum
amounts of data that can
be ingested by experts in the mixture.

Importantly, it can be seen that when $N >2$ there can be multiple instantiations
of $\boldsymbol b$ that  produce the same average data cost $\bar{d}$. Thus,
when an average data cost target $d_t\in [d_{min},d_{max}]$ is specified,
it is necessary to define a method by which to determine an appropriate bias
vector $\boldsymbol b$ that is subsequently used in training   biased mixtures.
To address this,  we consider  $\boldsymbol p$
where $p_n$ quantifies the  performance of each optimized expert prior to inclusion
in the mixture, and  select $\boldsymbol b$ such that: \textit{(i)} $\boldsymbol
b$ satisfies $\bar{d}=d_t $, and \textit{(ii)} $\boldsymbol b$ maximises
 the expected test performance as measured by $\boldsymbol b\boldsymbol p^\mathrm{T}$.
That is, when each
 component $p_n$ denotes an appropriate
performance measure for the $n^{th}$ expert on a designated  set of inputs
isolated from testing
examples (e.g.,
$ p_n$ can be accuracy
for classification tasks, or mean average precision for objection detection
tasks), $\boldsymbol b  \boldsymbol p^\mathrm{T}$ is a  measure
of  performance when examples are  assigned to experts with respect
to $\boldsymbol b$. In doing so, we reduce the problem of determining $\boldsymbol
b$ for a specified data
cost $d_t$ to a linear optimization problem that achieves  $\boldsymbol {b
d}^\mathrm{T}=d_t$, while maximising  $\boldsymbol {b p}^\mathrm{T}$.  Since
$||\boldsymbol b||_{_1}=1$ and $b_N$ can be expressed as $b_{N}= 1-\sum_{n=1}^{N-1}
b_{n}$,  by expanding and substituting $b_N$
we get the constraint:

\begin{equation}\label{eq:dt_constraint}
\begin{aligned}
b_1d_1+b_2d_2+...+ (1-\sum_{n=1}^{N-1} b_{n})d_N =d_t
\end{aligned}
\end{equation}and following that components of  $\boldsymbol b$ must be summable
to unity, we also get the additional $(N-1)$ constraints:

\begin{equation}\label{eq:less_than_one}
\begin{aligned}
b_1\leq 1 ; \; b_2 \leq 1 ; ...   \; ; \;   b_{N-1} \leq 1
\end{aligned}
\end{equation}with the performance maximization objective: 

\begin{equation}\label{eq:z_objective}
\begin{aligned}
\mathrm{max}\{b_1p_1+b_2p_2+...+b_Np_N\}
\end{aligned}
\end{equation}

Note that (\ref{eq:dt_constraint}) and (\ref{eq:less_than_one}) define $N$
linear constraints to maximize the objective (\ref{eq:z_objective}) with
$N$ basic values $\{b_1, b_2,..., b_N\}$. Following the duality property of such
convex problems  \cite{borgwardt2012simplex,fiedler2006linear}, we can also formulate the dual (and equivalent) problem that finds
$\boldsymbol b$ for any specified performance target $p_t$. That is, appropriate biases can be found to meet $p_t$ with the $(N-1)$ constraints of (\ref{eq:less_than_one}) and the additional constraint on expected performance:

\begin{equation}\label{eq:pt_constraint}
\begin{aligned}
b_1p_1+b_2p_2+...+ (1-\sum_{n=1}^{N-1} b_{n})p_N =p_t
\end{aligned}
\end{equation}
 with the data cost minimization objective:

\begin{equation}\label{eq:z_objective_pt}
\begin{aligned}
\mathrm{min}\{b_1d_1+b_2d_2+...+b_Nd_N\}
\end{aligned}
\end{equation}

Thus, determining $\boldsymbol
b$
is  a convex problem that can be
readily solved
by any convex optimization technique  \cite{borgwardt2012simplex,fiedler2006linear,boyd2004convex}, such as the simplex method \cite{borgwardt2012simplex, boyd2004convex}.
That is, an appropriate biasing value $\boldsymbol b$ to use for
training  can be found  for any
specified target data cost $d_t$ by solving for $\boldsymbol b$ in  (\ref{eq:dt_constraint})-(\ref{eq:z_objective}), or any target on expected performance $p_t$ by solving (\ref{eq:less_than_one}), (\ref{eq:pt_constraint}), and (\ref{eq:z_objective_pt}).

\subsection{Final Observations}
In considering the performance of biased mixtures, the quality of expert selections
from  $\mathcal{E}$ is  regulated by the complexity
of the gating function $G(\boldsymbol x; \mathcal{W}_g)$; where increasing the
complexity of $G(\boldsymbol x;\mathcal{W}_g)$ can improve selections (e.g., by increasing the number of learnable weights), albeit
with diminishing returns. In addition, and in the case of bias enforcement, we intuitively expect the quality of
selections to be directly correlated with  batch sizes used for training. That is,  
low batch size settings  may not expose gating functions to a sufficient amount
of variance in inputs to make  selections of benefit, and setting higher batch sizes  is favorable.

Importantly, applications of biased mixtures allow 
gating functions  $G(\boldsymbol x;\mathcal{W}_g)$ to wholly observe inputs $\boldsymbol x $ prior to selecting experts for data-economy. That is,  biased mixtures can be distributed such that they allow for gating before
preprocessing to produce
sampled inputs $P_n(\boldsymbol
x),$  and before inputs are subsequently sent to remote models for visual
inference (as illustrated in Figure \ref{fig:bmoe_illustration}). As a result, the constraint for gating functions
is not  input size,
but the processing capability on-board visual sensors. We also note that,  the expert selection methods detailed in Section \ref{sec:biased_expert_selection} can be applied on
mixtures comprising experts optimized via additional task-specific dimensionality reduction methods, and can also be applied on experts that use different modalities to make
their inferences (as illustrated in (c) of Figure \ref{fig:bmoe_illustration}). Finally, while our work studies the problem of reducing data utility, $\boldsymbol
b$ can also be specified to  prioritize any other expert  property whenever constraints are properly quantified and made available to the proposed gating architecture  (e.g., to meet constraints on power consumption or latency).

\section{Evaluation \\ \label{sec:evaluation}}

\subsection{Benchmarks and Evaluation Method \label{sec:evaluation_method}}

To show how biased mixtures
can optimize data costs of inference for different  problems, we evaluate
 on three computer vision tasks: \textit{(i)} object detection, \textit{(ii)}
image super resolution, and \textit{(iii)} realtime action classification.
In reporting results for all tasks, we compare our method against two alternatives:

\begin{enumerate}

\item  \textit{Previously Proposed Models}: To benchmark our results against
relevant  task-specific solutions, we consider the
performance
of   constituent
experts when optimized for different data cost constraints. In biased mixtures,
this corresponds to specifying  $\boldsymbol b$ as a one-hot vector, and
measures
performance
when the same amount of data is used for all inputs during inference (e.g.,
when $\boldsymbol b=[0,1,0]$ only $E_2$ is used for inference). We
report this to benchmark against previous work and  to highlight the benefit
of uniquely dividing the input
space for each 
expert.\ \ 

\item \textit{Random Selection}:  Here, experts are randomly selected for
inference
at test time in order to satisfy the model biasing requirement  $\boldsymbol
b$. This is to serve as the lower bound of performance when biased mixtures
are
used and the  specified expert utility bias is met.

\end{enumerate}

Importantly, when considering the problem of task-agnostic model optimization
under data cost constraints, there is no previous
work similar to ours (see Section \ref{sec:related_work}). That is why, we
benchmark against the maximum performance achievable
by recently proposed \textit{task-specific} solutions when their input
volumes are adjusted to meet different constraints on
data cost. That is, \textit{biased mixtures consist of
experts that also stand in as external benchmarks}.  To highlight the latter,
benchmark results of constituent experts are indicated in comparative plots
by markers on
 dotted lines.

For clarity,
and to ensure  consistency of  representation across all tasks, we report
 the per input  data cost of inference $\bar d$ as the average amount of
data seen by  the mixture   after inputs
are fully decompressed. For each evaluated task we specify how the data cost
 for each expert $d_n$ is measured (i.e., the data cost associated with $P_n(\boldsymbol
x)$).  For
a concise measure of how well models  preform across different specified
data cost constraints of
$d_t\in[d_{\min},d_{\max}]$,
 and with   $p_{\text{test}}(d_t)$
denoting  test performance  when 
the  target data cost is $d_t$, we report  the area under 
curve  when data cost is normalized as:

\begin{equation}\label{eq:sparseG}
\begin{aligned}
  \rho = \frac{}{}\int_{0_{}}^{1_{}}  p_{\text{test}}(d_{\min}+t(d_{\max}-d_{\min})) \;
dt
\end{aligned}
\end{equation}

For all mixtures, we specify the gating model (i.e., $f(\boldsymbol x; \mathcal{W}_g)$)
as a single  conv-pool layer  followed by a fully connected network. To ensure
that  the model selection process is of low
complexity for all tasks, we use
 ReLU activated depthwise separable convolutions \cite{sifre2014rigid}, and report the   per input number of multiply-accumulate
gating operations $C_g$. 
  We  use cross-validation to optimize the biasing weight $w_{\text{bias}}$
and report  the best performance when soft regularization is used. After
all experts included in the mixture are individually optimized,  biased
mixtures are trained by updating the weights of the gating function exclusively,
and the weights of experts are not fine-tuned further. We have  found that
using higher batch sizes is helpful when training biased mixtures, because
it exposes the mixture to a more varied set of input examples to partition
to each expert meaningfully. Therefore, to ensure gating functions learn
meaningful features for batch partitioning, for all tasks we set the batch
size to $128$ and the learning rate to $10^{-4}$.

\subsection{Single-Shot Object Detection}

We test our method on  single-shot detection (SSD) to reduce the data requirement
for object detection while maintaining
high accuracy.   Recent work  \cite{huang2017speed,howard2017mobilenets}
\cite{wang2018video}
  showed that SSD\
models vary widely in performance and complexity when input sizes are adjusted.
When considering the varying degrees of complexity of natural images, we
expect that the minimum required subsampling rate of inputs for accurate
object detection should vary accordingly. To demonstrate this, we
train a biased mixture of experts where each expert is optimized for a different
image subsampling rate, and use the recent work of Liu \textit{et. al}  \cite{liu2016ssd}
 as a baseline  for  all experts (for an illustration, see (a) of Figure
\ref{fig:bmoe_illustration}).   When the  resolution of   inputs to each
expert is $R_n\times R_n$ pixels, we measure the data cost associated with
$P_n(\boldsymbol x)$ as $3 \times R_n\times R_n\times K$,  where $3$ is the
number of color
channels in RGB inputs, and $K$ is the number of
bytes needed to store floating point decimals.

We use   VGG16      \cite{feichtenhofer2016convolutional} and ResNet50 
for feature extraction and evaluate all models using 300 regional proposal
boxes for VGG16 \cite{feichtenhofer2016convolutional}, and 50 regional proposal boxes for ResNet50 .
Following recent work \cite{huang2017speed,liu2016ssd}, we 
 train on COCO  training data while excluding the 8k mini-eval images used
in the 2012 challenge \cite{lin2014microsoft}, and  report performance as
the mean Average Precision (mAP)
on COCO  ($07$+1$2$). We  train  mixtures for $20$k steps to show our results when  using soft
regularization and bias enforcement, and in Table \ref{tab:ssd_c} we detail the
types and complexities of all layers used in devising the gating model $f(\boldsymbol x; \mathcal{W}_g)$.
 Inputs to the gating
model   are pre-processed as $224 \times224$ center crops of  $300 \times300$ images, 
 and we ensure
that the gating complexity of all mixtures remains at $C_g < 10^8$ Mult-Add
operations.

\begin{table}[h] \caption{ Single shot detection  comparison on COCO  \cite{lin2014microsoft}
of
biased mixtures  of  SSD \cite{liu2016ssd}
 experts against other benchmarks.   Resolutions \{$R_n$\} and
data costs \{$d_n$\} are reported for all experts.}
\centering
 \hspace*{-.6mm}

 \scalebox{0.9}{\begin{tabular}{|c|c|c|c|c|c|}

\multicolumn{6}{c}{$\{R_n\}=\{100, 150,300\}$(Pixels);  $\{d_n\}=\{120, 270, 1080\}$ (kB)}\\ 
\hline
\multirow{1}{*}{Feature} & \multirow{2}{*}{Biasing Method}  & \multicolumn{3}{|c|}{mAP$( d_t)$  ($\%$) when $d_t$  $=$} & \multirow{2}{*}{$\rho$} \\

Extractor & & \multicolumn{3}{|c|}{$d_{max}$ \;\; \; \small{$\frac{d_{max}}{2}$}
\;\;\;\;
\small{$\frac{d_{max}}{3}$}} & \\  

\hline

\multirow{4}{*}{ VGG16 }    & Benchmark Experts 
&\multirow{4}{*}{80.0}
& 70.0& 66.7& 70.9  \\

& Proposed $\boldsymbol b$ Enforcement  & & \textbf{72.5} &\textbf{70.9}&
\textbf{73.1}  \\

  & Soft Regularization \cite{shazeer2017outrageously}  & & 67.1& 65.0& 68.9
\\

& Random Selection  &  & 66.3& 63.4& 68.2  \\

\hline

\multirow{4}{*}{ResNet50  }    & Benchmark Experts  
&\multirow{4}{*}{75.7}
& 65.1& 61.3& 66.1  \\

& Proposed $\boldsymbol b$ Enforcement  & & \textbf{67.8}&\textbf{65.9}&
\textbf{68.3}  \\

  & Soft Regularization \cite{shazeer2017outrageously} & & 62.2& 59.9& 64.2
\\

& Random Selection  &  & 61.9& 57.4& 63.3  \\
\hline

 \multicolumn{2}{c}{} &  \multicolumn{1}{c}{ \textcolor[rgb]{1,1,1}{--------}}&
\multicolumn{1}{c}{\textcolor[rgb]{1,1,1}{----------}} &  \multicolumn{2}{c}{}\\

 \end{tabular}}   \label{tab:auc_ssd}
\vspace*{-3.5mm}  \end{table}

\begin{figure}[h] \vspace*{-3.5mm} \centering \centering \hspace*{-5.2mm}
\includegraphics[scale=0.22]{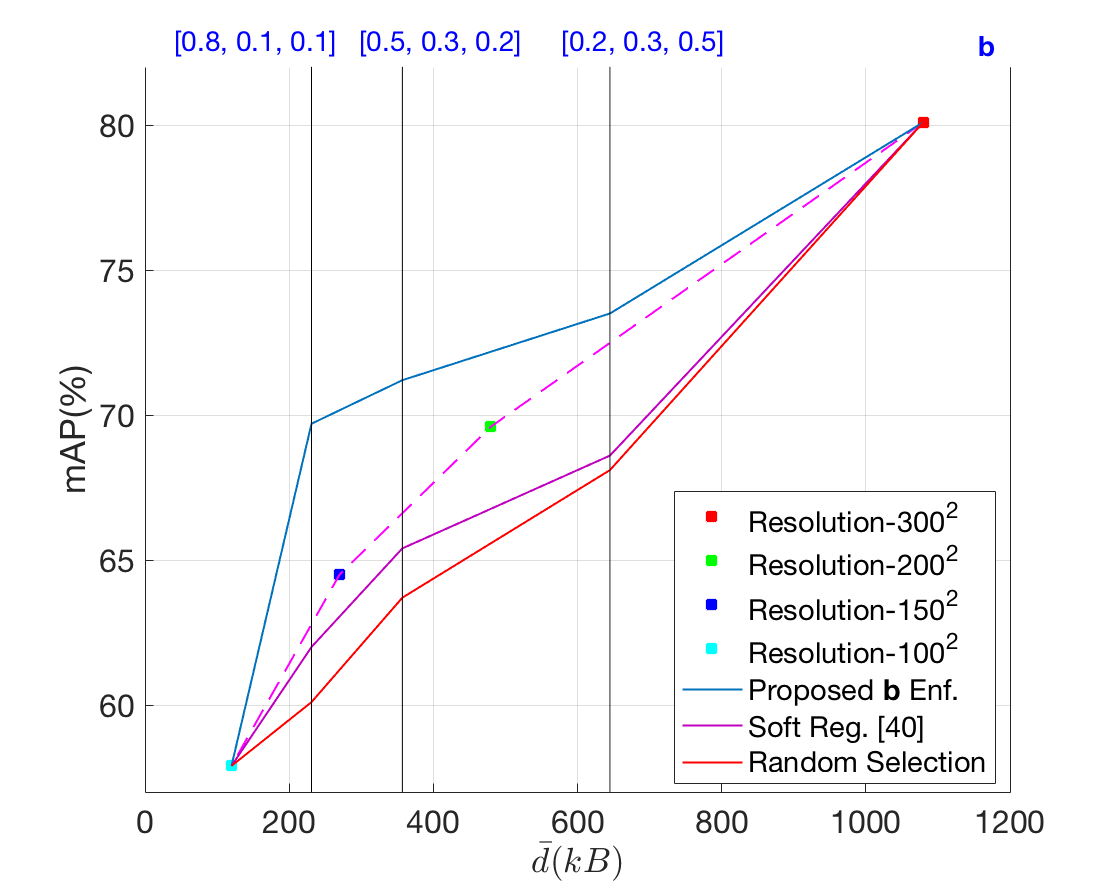}
      \vspace*{-1.5mm}     \caption{\label{fig:ssd} Single shot detection
performance comparison
of biased mixtures  of   VGG16 
experts against other benchmarks when   $\{R_n\}=\{100,150,300\}$.  The performance
of individual experts is shown on the dotted line. \ \ \ \ \    
\textbf{\textcolor{red}{}}} \vspace*{-1.5mm} \end{figure}

Figure \ref{fig:ssd} shows the relationship between imposed bias, data cost,
and mAP when three VGG16 experts are used for single shot detection, where
the resolution of inputs to each expert is $\{R_n\}=\{100,150,300\}$. Notably,
biased mixtures optimized with bias enforcement provide the slowest degradation
in mAP for lower data costs, with diminishing gains
when more data is available at test time. Specifically,  biasing via enforcement
outperforms    individual
experts by $7.5\% $ when an average of  $220$ kilobytes per image is allowed,
which is equal to the performance of individual experts
at $490$ kilobytes. That is, when the minimum acceptable mAP is  $70\%$,
a reduction of $270$ kilobytes in required data   is achieved by our proposal
(which is
equivalent to a saving of $55\%$ in bitrate).

In Table \ref{tab:auc_ssd} we
show the performance of biased mixtures when applied to multiple models,
 and report $\rho$  as a comprehensive measure of model performance across
data costs.  When compared to random selection, we note that for
both  ResNet50  and VGG16 , imposing
bias on  mixtures provides the highest gain when lower values of data cost
are considered
(e.g., when $d_t<\frac{d_{max}}{3}$).
Compared to soft regularization, and for all  mixture configurations, 
we found that bias enforcement is  a much more effective method for training
biased mixtures (this is also true for all  other tasks evaluated). We hypothesise
  this is because, when bias enforcement
is used only the task loss
is back-propagated during training, which causes less competition between
losses and therefore
less local minima to exist in solution space.

To further assess how biased mixtures learn useful bifurcations of input space, Table \ref{tab:ssd_expertwise} details the performance of each expert on their assigned subset of inputs. Notably, Table V  highlights how easier input examples are passed to the data-efficient expert $E_1$, resulting in  increased accuracies of $E_1$ compared to its baseline accuracy $57.91\%$, which is measured over all test inputs of COCO     \cite{lin2014microsoft}. Conversely, biased mixtures pass more difficult examples to $E_2$ and $E_3$, resulting in lower accuracies over their assigned inputs compared to their baseline accuracies.  Interestingly, and especially for bias enforcement, Table V also shows how improved  accuracies of $E_1$  (which correlate with how "easy" its assigned inputs are to classify) are inversely proportional to the number of examples passed to it, as reflected by $b_1$ (e.g., a  difference of $+12.72$ percentile points in accuracy when $b_1=0.8$, compared to an increase of $+18.31$ when $b_1=0.5$).

\begin{table}[t]
  \caption{ Relation between gating complexity, batch size, and performance
when bias enforcement is used. \textcolor{red}{}}
\centering
 \hspace*{-1.5mm}
\begin{footnotesize}
 \scalebox{0.9}{\begin{tabular}{|c|c|cccc|}

\cline{1-6}

\multirow{3}{*}{$C_g$ }  & \multirow{3}{*}{$M$} &  \multicolumn{4}{|c|}{ $ \rho$  when  $\{R_n\} = $}
\\
& \multirow{2}{*}{--} & \multicolumn{2}{|c|}{$\{100,300\}$(Pixels)} &\multicolumn{2}{|c|}{  $\{100, 150, 300\}$(Pixels) } \\  


(Mult-Adds) &&VGG16    & ResNet
\ &  \text{VGG16 }   &\ ResNet50

\\
\hline

\multirow{3}{*}{ 23,048,576} & 16 & 68.40 & 64.11 & 69.27 & 64.46
  \\

&32&  70.35& 65.89 & 70.25 & 65.57\\

 &64 &  \textbf{70.93}& \textbf{66.24} & \textbf{71.16}& \textbf{65.72} \
\\

\hline
    \multirow{3}{*}{26,194,304} &16 &70.85&  66.92 & 71.82 & 67.04 \\
&32 & 71.49 & 67.25 & 72.50& 67.41 \\
&64& \textbf{71.84}& \textbf{67.59} &\textbf{72.97}& \textbf{68.04}\\

\hline

   \multirow{3}{*}{38,700,216}   & 16& 70.93 & 67.01 &  72.10&  67.33 \\

& 32 &71.58 & 67.25& 73.07& 68.26\\

  & 64 &\textbf{71.86}&  \textbf{67.62} & \textbf{73.13} & \textbf{68.30}
\ 
\\
  \hline

\multicolumn{1}{c}{} & \multicolumn{1}{c}{\textcolor[rgb]{1,1,1}{---}} &
\textcolor[rgb]{1,1,1}{----} & \textcolor[rgb]{1,1,1}{-------} &\textcolor[rgb]{1,1,1}{----------}&
\multicolumn{1}{c}{\textcolor[rgb]{1,1,1}{------}}\\

 \end{tabular}} \end{footnotesize}
 \vspace{-5mm} \label{tab:c_m_rho} \end{table}

\begin{table}[h] \caption{  Layer complexities $C$ of the gating model $f(\boldsymbol
x;\mathcal{W}_g)$ for biased mixtures evaluated on single shot detection.
Expert input resolutions  are specified as $\{R_n\}=\{100,150, 300\}$ and
$N=3$.} 
\centering
 \hspace*{-2mm}
\begin{footnotesize}
\scalebox{0.9}{\begin{tabular}{|c|c|c|c|c|}

\hline

\multirow{2}{*}{Layer Type} & \multirow{2}{*}{Filter Shape}

& \multirow{2}{*}{Stride} &   \multirow{2}{*}{Input Shape} & \multirow{1}{*}{$C$}
\\

 & & & & (Mult-Adds)\\  

\hline

Convolutional & $3\times3\times3\times64$ & $2$ & $224\times224\times3$&
$2,747,136$ \\

Avg. Pooling & $7\times7$ & $5$ & $111\times111\times64$& ---- \\

Flatten Op. & $-$ & $-$ & $21\times21\times64$& ---- \\

Fully Connected & $28224\times1024$  & $-$ & $1\times 28224 $&$28,901,376$
\\

Fully Connected & $1024\times 3$ & $-$ & $1\times 1024$& $3072 $
\\

\hline

 \end{tabular}} \end{footnotesize}   \label{tab:ssd_c}
\vspace*{-2mm}  \end{table}

In Table \ref{tab:c_m_rho} we study the effect of adjusting the  gating complexity
$C_g$, batch size $M$, and number of experts $N$ on the performance of biased
mixtures when bias enforcement is used. When we consider all mixtures, we
find that batch size is critical to  performance. This is because bias is
enforced on a per batch basis, and to make meaningful decisions the gating
function needs to be exposed to an ample amount of variance between
examples. We also see that increasing the complexity of  gating  does increase
performance by helping partition the input space more effectively. However,
this effect saturates at $C_g\approx3.8\times10^7 $ Mult-Add operations,
which demonstrates that the optimal hyperplane  to partition input space
for  $N\leq3$ experts can be learned with  low complexity. 

\begin{figure}[h] \hspace*{-1mm}\includegraphics[scale=0.25]{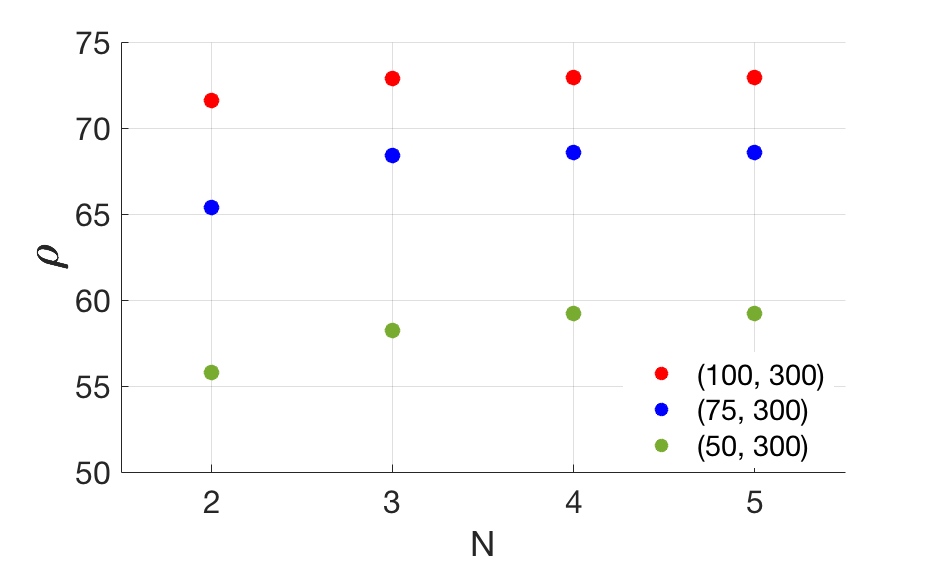}
\vspace{-8pt} \caption{  $\rho$ when  bias enforcement is used and the number
of experts $N$ is configured.   VGG16  is used for
feature extraction, and different colors indicate  the resolution  limits
 $(R_{min},R_{max})$ allowed to the mixture (where $N$ determines the number
of intermediate input resolutions included).}\label{fig:ssd_n_auc} \vspace{-4pt}
\vspace*{-.5mm} \end{figure}

By comparing the left and right part of  Table \ref{tab:c_m_rho}, we see
 that adding more experts to the mixture provides a modest
increase to performance.
This is because having more experts allows the mixture
to further exploit the variance in different input sub-spaces (if any such
variance exists). To see the extent to which this is
true, in Figure  \ref{fig:ssd_n_auc}   we  adjust the limits  of allowed
input resolutions
to the mixture  $R_{min}$ and $R_{max}$, and report $\rho$ when considering
different values of $N$.    Importantly, we see that when the difference
between $R_{min}$ and $R_{max}$ is lower, using more experts yields less
gain in performance, to the point where using more than three experts for
$(R_{min}, R_{max})=(100,300)$ does not provide any benefit. This is because,
while setting high values of $N$ increases the number of intermediate resolutions
between $R_{min}$ and $R_{max}$,   the difference $(R_{max}-R_{min}) $ correlates with the amount
of discernable adequacy between experts, which in turn correlates with 
the benefit of including more experts.

\subsection{Image Super-Resolution}

We test the applicability of biased mixtures on Single Image Super resolution
(SISR), an image reconstruction task where spatial features of high-resolution
images are inferred from low-resolution input images. Several recent   proposals
 have  shown good performance in terms of  image reconstruction
accuracy and computational efficiency \cite{shi2016real,dong2016accelerating,timofte2017ntire}\cite{zhang2018single}.
However, current super resolution models do not take into account
the variable amount of high-frequency edge content between images. That is,
 when reconstructing images
which contain many high frequency elements, SISR models are likely to benefit
from higher
resolution input images, while images comprising predominately low-frequency
content can be inferred just as well from lower resolution inputs.
This is true also when considering different parts of an image, which usually
vary in  the breadth of their frequency elements. To demonstrate this, we   train biased mixtures   to
determine the needed input resolution
  for   good image reconstruction, and in doing so, we show how different image parts can be adaptively upsampled subject to their content. Such decisions about selected super-resolution experts can also be augmented to existing media streaming standards (e.g., DASH/HLS in HTTP
[19]) for adaptive
subsampling prior to transmission.

We evaluate on the NTIRE17 challenge
dataset DIV2K  \cite{agustsson2017ntire}, and use state-of-the-art proposals on super-resolution \cite{shi2016real,dong2016accelerating} as baselines for constituent experts of biased mixtures.  To expose 
biased mixtures   to the  intra-image variance of frequency elements,  images
are divided using a fixed grid into parts of size $64\times 64$\ pixels,
and super-resolution
is performed on each part separately (for an illustration, see (b) of Figure
\ref{fig:bmoe_illustration}). By inspecting the  low-level semantics of 
each image part, the mixture selects the
most data efficient expert for reconstruction to preform an upscaling from
the set  $\{S_n\}=\{\times4,\times3,\times2\}$. For each expert that
upscales inputs with a factor of $S_n$ to match the target resolution of
$64\times 64$ pixels, we measure the associated   data cost as $d_n=(64/S_n)^2
\times K$, where $K$ is the number of bytes needed to store floating point
decimals. To expose gating
to the high frequency components of input images, inputs to the
gating model  are not subsampled, and are maintained at the original resolution of resolution of   $64 \times 64$ pixels.
For all biased mixture results, mixtures are trained for $20$
epochs and we ensure the complexity of the gating function is set to $C_g<10^{7}$
Mult-Add operations.

 \begin{figure}[h] \centering \centering \hspace*{-4.9mm} \vspace*{-.5mm}\includegraphics[scale=0.25]{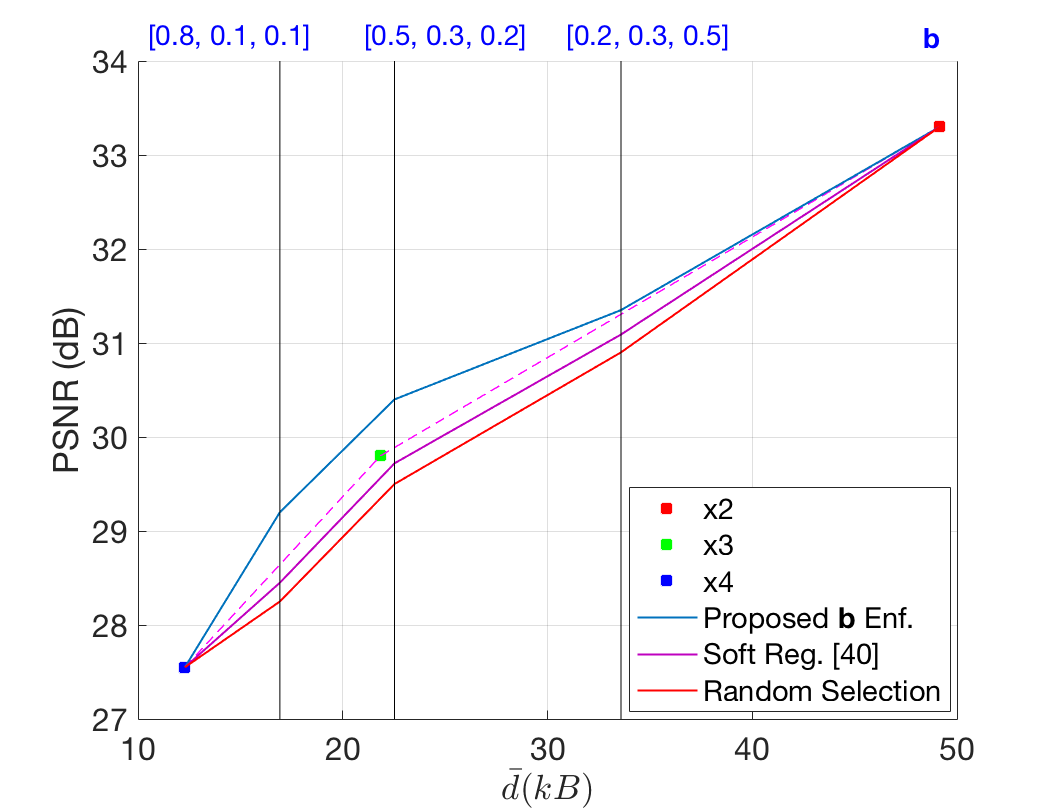}
       \caption{\label{fig:super_resolution} 
Super resolution performance
comparison of biased mixture of ESPCN \cite{shi2016real} experts and other
benchmarks when \{$S_n$\}$=\{\times 4, \times 3,\times 2\}$. \    
\textbf{\textcolor{red}{}}}  \vspace*{-.mm} \end{figure}

\begin{table}[h] \caption{ Image super resolution comparison on DIV2K  \cite{agustsson2017ntire}
of
biased mixtures and other benchmarks.  Upscale factors \{$S_n$\} and
data costs \{$d_n$\} are reported for all experts.}
\centering
 \hspace*{-0.7mm} \vspace*{-0.7mm}
\begin{footnotesize}
\scalebox{0.9}{\begin{tabular}{|c|c|c|c|c|c|}

 \multicolumn{6}{c}{\{$S_n$\}$=\{\times 4, \times 3,\times 2\}$;  $\{d_n\}=\{13.9, 21.8, 49.2\}$ (kB)}\\ 
\hline

\multirow{2}{*}{Model} & \multirow{2}{*}{Biasing Method} & \multicolumn{3}{|c|}{PSNR (dB) when $d_t=$}& \multirow{2}{*}{$\rho$} \\

 & & \multicolumn{3}{|c|}{$d_{max}$ \;\; \; \small{$\frac{d_{max}}{2}$} \;\;\;\;
\small{$\frac{d_{max}}{3}$}} & \\  

\hline

\multirow{4}{*}{ESPCN\cite{shi2016real} }    & Benchmark Experts \cite{shi2016real}

&\multirow{4}{*}{33.3}
& 30.4& 28.4& 30.7  \\

& Proposed $\boldsymbol b$ Enforcement  & & \textbf{30.7} &\textbf{28.8}&
\textbf{31.0}  \\

  & Soft Regularization \cite{shazeer2017outrageously} & & 30.0& 28.1& 30.6
\\

& Random Selection  &  & 29.8& 28.0& 30.5  \\

\hline

\multirow{4}{*}{F-SRCNN \cite{dong2016accelerating} }    & Benchmark Experts
 \cite{dong2016accelerating}
&\multirow{4}{*}{32.8}
& 29.8& 28.0& 30.3  \\

& Proposed $\boldsymbol b$ Enforcement  & & \textbf{30.1}&\textbf{28.3}&
\textbf{30.5}  \\

  & Soft Regularization \cite{shazeer2017outrageously} & & 29.3& 27.6& 30.1
\\

& Random Selection  &  & 29.2& 27.5& 30.0  \\

\hline

\hline

 \multicolumn{2}{c}{} &  \multicolumn{1}{c}{ \textcolor[rgb]{1,1,1}{--------}}&
\multicolumn{1}{c}{\textcolor[rgb]{1,1,1}{----------}} &  \multicolumn{2}{c}{}\\

 \end{tabular}} \end{footnotesize}   \label{tab:auc_sr}
\vspace*{-0.1mm} \end{table}

In Table \ref{tab:auc_sr} we compare biased mixtures   against other benchmarks
when using ESPCN\ \cite{shi2016real} and FRSCNN \cite{dong2016accelerating}
as baselines, in Table
VI we detail the performance  of experts over their
assigned subsets of input, and  in Figure \ref{fig:super_resolution}
we show the relationship between average data cost and PSNR when considering
ESPCN\ \cite{shi2016real}.  Notably from Figure \ref{fig:super_resolution},
  when bias enforcement is used and $\bar d$ is within the range
of $18\text{-}22$ kilobytes, biased mixtures  outperform single experts with
an average difference of $0.4$ dB. Over the same range of values of $\bar d$,   and when compared to random selection, bias enforcement provides an
average improvement of $0.7$ dB. This highlights the magnitude of intra-image
high
variance in required input resolution for image reconstruction, which is
not considered by  random selection and  optimized experts. 
Overall, Figure \ref{fig:super_resolution} and Table \ref{tab:auc_sr}   show
that    biased mixtures
outperform individual experts most when $\bar d<20$ kilobytes, with diminishing
gains in performance for higher values of $\bar d$.  Consistent with our observations on object detection, Table VI shows how easier inputs  are passed to the data-efficient super-resolution model $E_1$, thereby increasing its reconstruction accuracy, while more difficult examples are passed to $E_2$ and $E_3$ resulting in a  modest reduction of their PSNR\ performance.

\vspace*{-1mm}

 \begin{figure}[h] \vspace*{-0.1mm} \hspace*{-1.5mm} \includegraphics[scale=0.5]{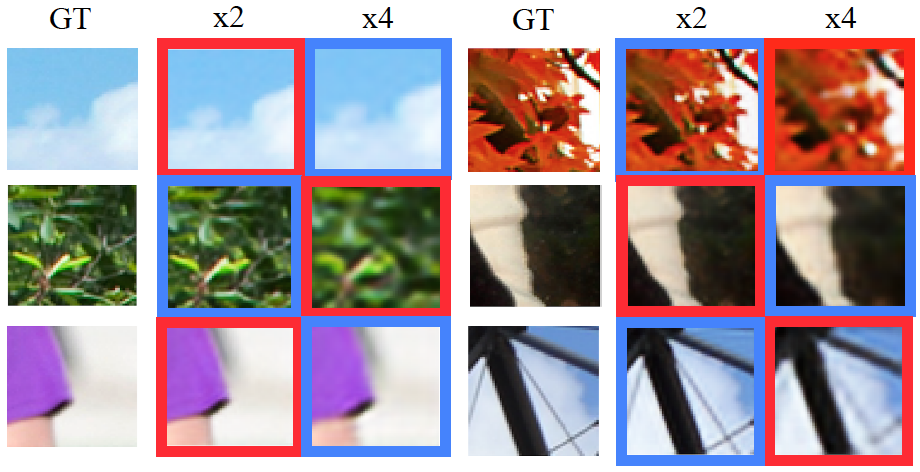}
   \vspace*{-2mm}       \caption{ Examples of expert assignments to different
image parts. Selected and non-selected experts are respectively highlighted
by  blue and red borders. Note the exploitable variance in detail between
images, which translates into the data cost savings reported in Table \ref{tab:auc_sr}.
}  \label{fig:super_resolution_visual_}\vspace*{-4mm}\end{figure}

\begin{table}[h] \caption{  Layer complexities $C$ of the gating model $f(\boldsymbol
x;\mathcal{W}_g)$ for biased mixtures evaluated on single image super-resolution.
 Expert upscaling factors are specified as $\{S_n\}=\{\times 4,\times 3,
\times 2\}$ and $N=3$.}
\centering
 \hspace*{-2.3mm}
\begin{footnotesize}
\scalebox{0.9}{\begin{tabular}{|c|c|c|c|c|}

\hline

\multirow{2}{*}{Layer Type} & \multirow{2}{*}{Filter Shape}

& \multirow{2}{*}{Stride} &   \multirow{2}{*}{Input Shape} & \multirow{1}{*}{$C$}
\\

 & & & & (Mult-Adds)\\  

\hline

Convolutional & $3\times 3\times3\times64$ & $2$ & $64\times64\times3$&
$224,256$ \\

Avg. Pooling & $3\times3$ & $2$ & $21\times21\times64$& ---- \\

Flatten Op. & $-$ & $-$ & $10\times10\times64$& ---- \\

Fully Connected & $6400\times512$  & $-$ & $1\times 6400 $&$3,276,800$\\

Fully Connected & $512\times 3$ & $-$ & $1\times512$&$1,536$$$
\\

\hline

 \end{tabular}} \end{footnotesize}   \label{tab:sr_c}
 \vspace*{-0.1mm}  \end{table}

In Figure \ref{fig:super_resolution_visual_} we  show  examples of  expert
selections made by the biased mixture to resolve  different $64\times 64$
inputs when bias enforcement is used.   The mixture  learns to pass image
parts with  high frequency components to the $\times2$ SISR model,  and passes
other less demanding parts to the  $\times4$ model (which  are blurrier,
due to the lower frequency of their components).

\subsection{Realtime Action Classification\\ \\ }

\vspace*{-3mm}

We   validate biased expert selection on \textit{realtime}  video action classification  in the
compressed  domain. While the best performing action classification models
operate on   uncompressed
video data, 
to reduce latency, the models proposed in recent work 
approximate a  low-resolution optical flow
from codec motion vectors at high speeds for action classification. The
 classifiers of 
 use two-stream architectures to infer actions,
 where    spatial and temporal classifiers complement each other by learning
different sets of features from their respective domains.
As such, for some action subsets, the use of only the spatio-temporal or spatial
classifier can  suffice
in drawing accurate distinctions between actions, but combining the predictions
of both  yields the highest accuracy.

Distinct from other compute-exhaustive models for action classification,   recent proposals on \textit{realtime} video classification 
use  minimal volumes
of data  to ensure complexities and runtimes remain low. Moreover, and in order to bypass complexity overheads associated with dense optical flow
estimation, the proposals of produce spatio-temporal modes 
 (which we hereon refer to as temporal modes for brevity) directly from compressed bitstreams. 
To show how biased mixtures can be applied in multi-modal settings, we expose   temporal modes of video to gating functions that appropriately select  which modes are used for subsequent classification. That is, \textit{prior} to sending inputs to remote  realtime classification models, we show how gating functions of biased mixtures can opt to use modalities only when they are needed for accurate classification (and we illustrate this
in  (c) of Figure \ref{fig:bmoe_illustration}). By doing so, we show how biased mixtures can  learn to leverage modal variance to mitigate unnecessary
traffic between sensors and remote  classifiers. 

We  evaluate on UCF-101 \cite{soomro2012ucf101} 
and measure the cost associated with the spatial mode as  $F\times H_s\times
H_{s}\times W_s\times K\times3$, where $F_s=2$ is the number of RGB frames used, $H_s=360$
and $W_s=240$ are the height and width of  inputs, $3$ is the number of color channels, and $K$ is the number
of bytes to store floating point decimals. For the temporal model, we measure
the data cost as $F_t\times H_t\times W_t\times K\times2$, where $F_t=150$
is the number of frames used,   $H_t=24$
and $W_t=24$ are the height and width of motion vector maps,    and $2$ is the number of channels used to represent
vertical and horizontal motion.
Importantly, we select spatial sampling rates akin to those of\cite{chadha2017video} which sets $F_{s}=1$,   $F_t \geq 10$, and the proposal of  \cite{zhang2018real} which sets $F_{s}=1$,  $F_t \geq 100$. This is to meet complexity limits for realtime inference, where the benchmark models set modest sampling rates compared to other exhaustive methods \cite{carreira2017quo}, which typically use  dense optical flow approximations with  $F_s\geq25$ and $F_t\geq250$. Moreover,  in implementing the benchmark model of Zhang \textit{et al.} \cite{zhang2018real}, we follow their method of upsampling  $24\times24$ motion vector maps  to $224\times224$ temporal mode inputs, and we specifically use a nearest-neighbours upsampling filter. Inputs are upsampled after  they are sent via the $(*)$ operator of Figure 2 (c), and therefore input shape parameters remain at $H_t=24$
and $W_t=24$ when measuring data cost.

For data-exhaustive action classification, we use  fusion classifiers  which combine both modalities
to predict actions with the highest possible accuracy. Fusion classifiers incur a data cost equal
the sum of both modalities. We include all modalities to train biased mixtures
of  experts, where $\{\text{Mode}_n\}=\{\text{Temporal},\text{Spatial},\text{Fusion}\}$. Importantly, and to allow for lower complexities of gating, inputs to the
gating model include only the temporal modes of   videos,  and spatial modes are not used.
For all biased mixtures, we train for $80$k steps and restrict the
complexity of the gating function
to $C_g < 10^8$ Mult-Add operations, where we detail the layer-wise complexities of gating in Table \ref{tab:rtac_c}.

\begin{table}[h] \caption{ Realtime action classification on UCF-101\cite{soomro2012ucf101}
of
biased mixtures of experts and other benchmarks.  Modalities  \{$\text{Mode}_n$\}
and
data costs \{$d_n$\} are reported for all experts.}
\centering
 \hspace*{-2.5mm}
\begin{footnotesize}
\scalebox{0.9}{\begin{tabular}{|c|c|c|c|c|c|} 
%
\hline

\multirow{2}{*}{Model} & \multirow{2}{*}{Biasing Method}  & \multicolumn{3}{|c|}{Accuracy$(d_t)$ (\%) when $d_t$ $=$} &   \multirow{2}{*}{$\rho$} \\

 & & \multicolumn{3}{|c|}{$d_{max}$ \;\; \; \small{$\frac{d_{max}}{2}$} \;\;\;\;
\small{$\frac{d_{max}}{3}$}} & \\  

\hline

\multirow{4}{*}{MV-3DCNN\cite{chadha2017video}}    & Benchmark Experts \cite{chadha2017video}&\multirow{4}{*}{88.0}
& 79.0& 77.9& 80.9  \\

& Proposed $\boldsymbol b$ Enforcement  & & \textbf{82.0} &\textbf{80.4}&
\textbf{83.5}  \\

  & Soft Regularization \cite{shazeer2017outrageously} & & 80.3& 78.0& 81.9
\\

& Random Selection  &  & 78.8& 77.3& 81.3  \\

\hline

\multirow{4}{*}{EMV-CNN \cite{zhang2018real}}    & Benchmark Experts \cite{zhang2018real}
&\multirow{4}{*}{85.6}
& 76.6& 75.5& 78.7  \\

& Proposed $\boldsymbol b$ Enforcement  & & \textbf{80.2}&\textbf{79.2}&
\textbf{81.3}  \\

  & Soft Regularization \cite{shazeer2017outrageously} & & 77.2& 75.6& 79.7
\\

& Random Selection  &  & 75.7& 74.9& 79.0  \\

\hline

 \multicolumn{2}{c}{} &  \multicolumn{1}{c}{ \textcolor[rgb]{1,1,1}{--------}}&
\multicolumn{1}{c}{\textcolor[rgb]{1,1,1}{----------}} &  \multicolumn{2}{c}{}\\

 \end{tabular}} \end{footnotesize}   \label{tab:auc_rtac}
\vspace*{-7mm}  \end{table}

\begin{table}[H] \caption{  Layer complexities $C$ of the gating model $f(\boldsymbol
x;\mathcal{W}_g)$ for biased mixtures evaluated on realtime action classification.
 Expert modalities  are specified as $\{\text{Mode}_n\}=\{\text{Temporal},\text{Spatial},
\text{Fusion}\}$ and $N=3$. Note that the gating model  $f(\boldsymbol
x;\mathcal{W}_g)$ ingests only temporal modalities of $\boldsymbol x$. }
\centering
 \hspace*{-2.5mm}
\begin{footnotesize}
\scalebox{0.9}{\begin{tabular}{|c|c|c|c|c|}

\hline

\multirow{2}{*}{Layer Type} & \multirow{2}{*}{Filter Shape}

& \multirow{2}{*}{Stride} &   \multirow{2}{*}{Input Shape} & \multirow{1}{*}{$C$}
\\

 & & & & (Mult-Adds)\\  

\hline

Convolutional & $3\times 3\times 320\times64$ & $2$ & $24\times24\times320$&
$3,363,840$ \\

Flatten\ Op. & $-$ & $-$ & $11\times11\times64$& ---- \\

Fully Connected & $7744\times1024$  & $-$ & $1 \times 7744 $&$7,929,856$
\\

Fully Connected & $1024\times 3$ & $-$ & $1\times1024$&$3,072$
\\

\hline
 
\end{tabular}} \end{footnotesize}   \label{tab:rtac_c}
\end{table}

In Table \ref{tab:auc_rtac} we compare the performance  of biased
mixtures against other benchmarks when using  the spatial and temporal classifiers
of  \cite{chadha2017video} and  \cite{zhang2018real} as baselines, and in Table
VII we detail the performance  of experts over their
assigned input subsets as determined by $G(\boldsymbol x;\mathcal{W}_g)$.
From Table \ref{tab:auc_rtac}, we first note that both biasing methods outperform random selection, by
up to $1\%$ for soft regularization and up to $3.8\%$ for bias enforcement.
This indicates that the biased mixture learns to discern confusing classes
for particular modalities to pass them to others. Notably, when
 $\bar d=\frac{d_{max}}{3}=860$ kilobytes, bias enforcement gives an accuracy
$1.4\%$ higher than that of the optimized experts at $\frac{d_{max}}{2}=1290$
  kilobytes, which requires  $430$ kilobytes more in data cost.

\begin{figure}[h] \vspace*{-0.1mm} \centering \hspace*{-3.mm}     \includegraphics[scale=0.21]{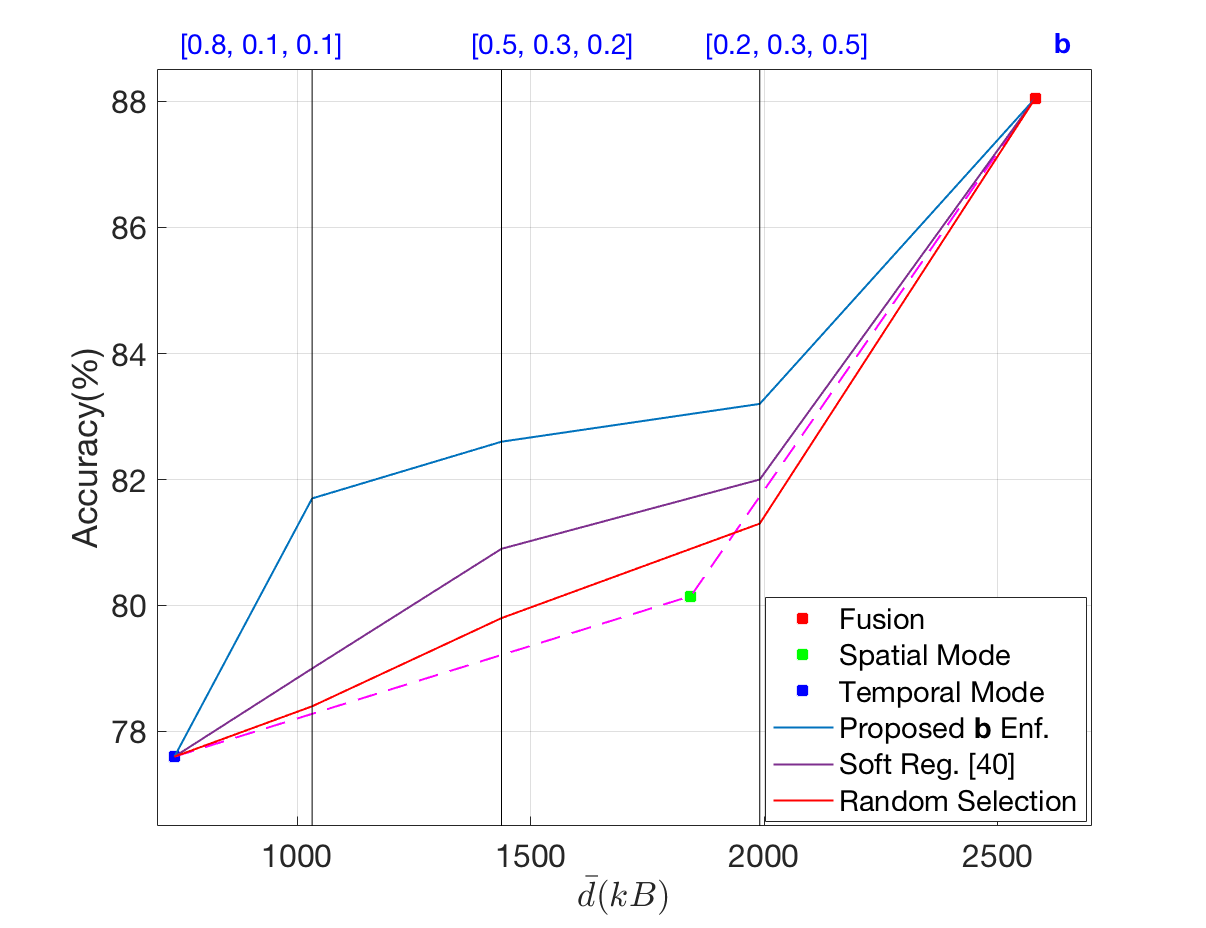}
 \vspace*{-1mm}\caption{\label{fig:action_classification_fusion_mv} Realtime
action classification performance comparison  of biased mixtures of MV-3DCNN
\cite{chadha2017video} experts,
with expert modalities   \{$\text{Mode}_n$\}=$\{$Temporal,Spatial, Fusion$\}.$
}\vspace*{-4.5mm}  \end{figure}

In Figure \ref{fig:action_classification_fusion_mv} we show the relationship
between $\bar d$ and action classification accuracy for instances of $\boldsymbol
b$ when 
 biased mixtures  of MV-3DCNN
\cite{chadha2017video} experts are used and the  mode of each expert is $\{\text{Mode}_n\}=\{\text{Temporal},
\text{Spatial}, \text{Fusion}\}$.
We first note that, due to the low resolution of its inputs, the temporal
classifier requires
the least amount of data and can predict actions with
an accuracy of $77.8\%$. By selecting among the three modes, both biasing
methods outperform random selection, with bias enforcement increasing accuracy
by up to $3.4\%$
for when $\bar d=1032$ kilobytes. Notably, and
when
using the temporal classifier for $80\%$ of  videos at $\bar d=1032$ kilobytes
(i.e., when $\boldsymbol b =[0.8,0.1,0.1]$),
bias enforcement is $1.6\%$ more accurate than the spatial
classifier (which requires  $811$ kilobytes more in data, equivalent
to an increase of $78\%$ in data cost).
The latter shows
 the extent to which biased mixtures can improve performance by using modest
amounts of data, even compared to individual models that require substantially
more in data cost.

Table VII shows how inputs are  appropriately passed to  experts for data-economic classification. Specifically, it shows how biased mixtures learn to use the data-efficient temporal model for inputs that are easier to classify, where  temporal modalities are likely to suffice for accurate classification.  For example, this is evident  when $b_1=0.5$ and $b_1=0.2$, where the temporal classifier $E_1$ respectively gains $+5.81$ and $+5.60$ percentile points in classifying its assigned inputs when compared to its baseline accuracy measured over  all test videos of UCF-101 \cite{soomro2012ucf101}.  On the other hand,   Table VII also shows how more difficult inputs are passed to the spatial and fusion classifiers, resulting in a modest  loss of accuracy  when classifying their assigned inputs. Moreover, Table VII  highlights how bias enforcement is superior to soft regularization in assigning inputs to different modalities, where this is evident across all values of $\boldsymbol b$. \\ 

\begin{figure}[h] \centering \centering \hspace*{-2.mm} \includegraphics[scale=0.2]{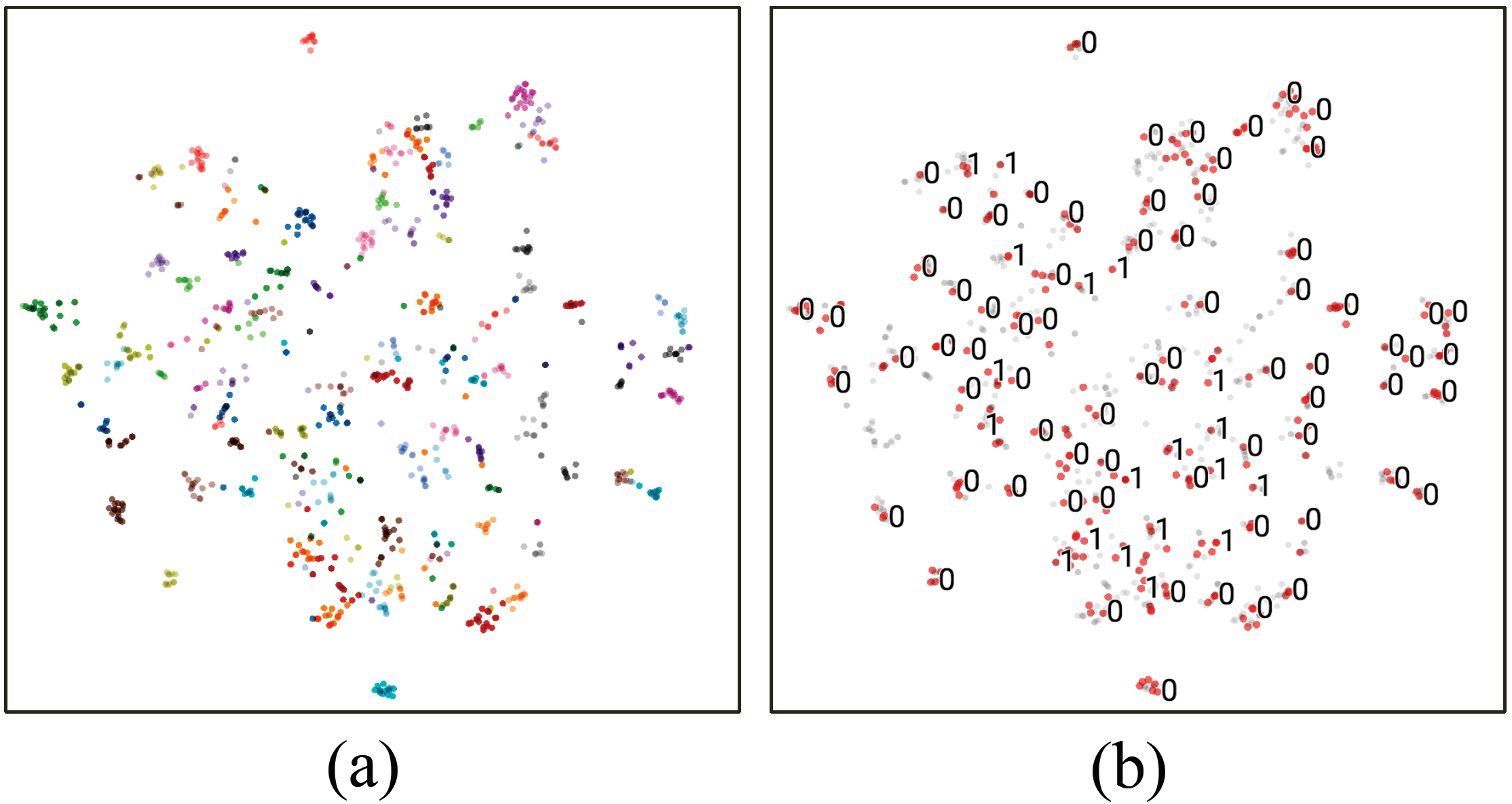}
          \caption{\label{fig:rtac_tsne_example} t-SNE \cite{maaten2008visualizing}
projections of $1024$ UCF101 videos, where  $\{\text{Mode}_n\}=\{\text{Temporal}, \text{Fusion}\}$
and $\boldsymbol
b =[0.75, 0.25]$. In (a)  
colours indicate different classes, and in (b)  mode assignments are shown as
$0$ or 1 for the temporal and  fusion classifiers respectively. \textit{Zoom in
to view  in high-resolution}.} 
\vspace*{-2mm} \end{figure}

\begin{figure}[h] \centering \vspace*{-8.mm} \centering \hspace*{1.mm}  \vspace*{-5.mm} \includegraphics[scale=0.25]{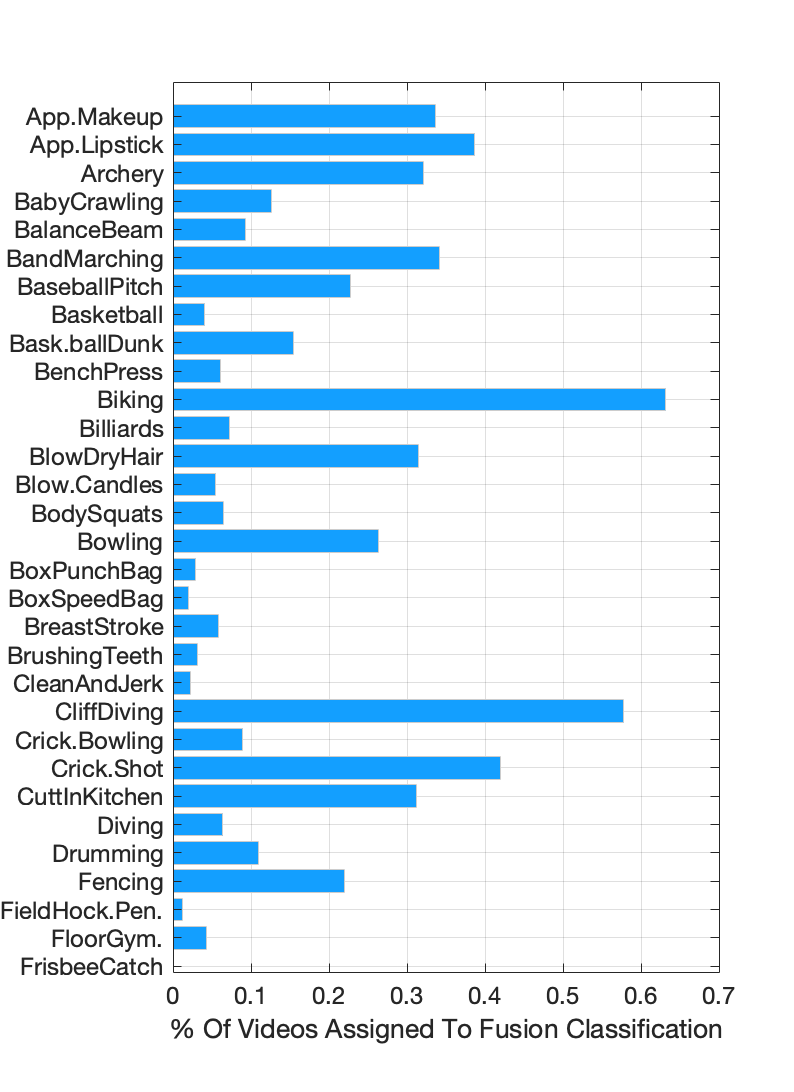}
          \caption{\label{fig:rtac_classes} Percentile of videos assigned to the the fusion classifier for the first $30$ classes of UCF101\cite{soomro2012ucf101}, where  $\{\text{Mode}_n\}=\{\text{Temporal}, \text{Fusion}\}$
and $\boldsymbol b =[0.75, 0.25]$.} 
\vspace*{-2mm} \end{figure}

To visualize how   different modalities are assigned to videos, in Figure
 \ref{fig:rtac_tsne_example} we show  two-dimensional t-SNE \cite{maaten2008visualizing}
 projections of $1024$  UCF101 examples as embedded by the last layer of
the  temporal classifier. We train via bias enforcement, and for clarity of presentation, we use a biased mixture
of two modalities $\{\text{Mode}_n\}=\{\text{Temporal}, \text{Fusion}\}$
where $\boldsymbol
b =[0.75, 0.25]$. In this way, we show the  relation between different  class
labels and assigned modalities.
In Figure 9 (a)  the middle region highlights instances of different classes which are more entangled and therefore harder to classify.  
For a sample of instances, Figure \ref{fig:rtac_tsne_example} (b) shows modalities selected by the biased mixture for action classification. Notably, the biased
mixture tends to select the data-exhaustive fusion mode for instances located in the entangled \textit{middle region},  where inputs are harder to classify (as indicated by label $1$ in (b) of Figure \ref{fig:rtac_tsne_example}),
and    temporal modes are  predominantly used for sufficiently 
isolated input clusters located \textit{outside the middle region} (as indicated by label $0$ in (b) of Figure
\ref{fig:rtac_tsne_example}). That is, Figure \ref{fig:rtac_tsne_example} shows how the biased mixture favors using the temporal classifier for video clusters that are comparatively isolated  and easy to discern, while the fusion model is used when videos are more entangled and harder to classify. 

For the same biased mixture that yields the t-SNE representation of Figure  \ref{fig:rtac_tsne_example}, in Figure \ref{fig:rtac_classes} we detail the classwise percentile of videos assigned to the fusion classifier. Evidently from Figure \ref{fig:rtac_classes}, challenging  inputs   are typically sent to the data-exhaustive fusion mode when they contain: (i) significant
camera movement, leading to noise in underlying  motion flow (e.g., for $63\%$ and $57\%$ of ``Biking" and ``Cliff Diving" instances, respectively),  and (ii) relatively static scenes,
resulting in sparse motion vector maps (e.g., for $38\%$ and $32\%$ of ``Apply Lipstick" and ``Blow Dry Hair" instances, respectively). 
 Hence, Figure 9 and Figure \ref{fig:rtac_classes} show
how biased expert  mixtures can find useful bifurcations of input space such
that  only necessary modalities are used for action classification, and less
data is used whenever possible.

\section{Conclusion \label{sec:conclusion}}

We introduce  biased expert utility in mixtures-of-experts  for effective partitioning
of input space to meet constraints on data availability at test time.  We propose two methods for training biased
mixtures, and evaluate their performance on multiple models for
all investigated   tasks. We
show how biased mixtures are applicable to any situation wherein experts
vary in data requirement and performance, and demonstrate this on a wide
range of computer vision tasks (we also make public a Tensorflow-$1.14$ implementation of biased mixtures in \texttt{https://github.com/UCL-Abbas/bmoe}).
Our validation shows that, especially for lower
ranges
of allowed data cost, biased mixtures  significantly outperform  baseline models
optimized to meet the same constraints on available data. We also show how useful gating inferences that prioritise data
economy can be realized with complexities that do not exceed $10^8$ Mult-Add
operations, which are feasible to run even on embedded
computation units (e.g., ARM Cortex-M7). Within contexts of  distributed visual inference, and to meet different
constraints on data transfer and bandwidth at test time, all of our observations
and tests show the importance of conditioning
 data utility for visual inference to the local proximities and properties
of inputs within their space. In other words, the importance of doing so
is applicable to all presented vision tasks, and is likely to extend to other
visual inference tasks in order to mitigate unnecessary burdens on communication
resources and sensor hardware.  We finally
note that an important advantage of biased mixtures is the flexibility at
which they
can be applied, in that, biased mixtures \textit{do not modify }their constituent
experts, but rather \textit{augment} their function with an input preprocessing
stage
that allows for data-economic inference.

\appendices


{\small
\bibliographystyle{IEEEtrans}
\bibliography{literature}
}



\end{document}